\documentclass[3p,11pt]{elsarticle}

\usepackage[utf8]{inputenc}
\usepackage{wrapfig}

\usepackage{makecell}

\usepackage{hyperref}
\usepackage{zref-clever}

\usepackage{float}

\usepackage{graphicx} 
\usepackage{amsmath}
\usepackage{amsfonts}
\usepackage{amssymb}
\usepackage{amsthm}
\usepackage{comment}

\usepackage{booktabs}
\usepackage[normalem]{ulem}
\usepackage{multirow}
\usepackage{enumitem}
\usepackage{dblfloatfix}
\usepackage{appendix}
\usepackage{xhfill}

\usepackage{pdfpages}
\PassOptionsToPackage{natbib=false}{biblatex}

\usepackage{pifont}

\makeatletter
\def\ps@pprintTitle{%
  \let\@oddhead\@empty
  \let\@evenhead\@empty
  \def\@oddfoot{\reset@font\hfil\thepage\hfil}
  \let\@evenfoot\@oddfoot
}
\makeatother

\newcommand{\E}{\mathbb{E}}
\newcommand{\Err}{\mathcal{E}}

\begin{document}

\begin{frontmatter}

\title{
Distilling Lightweight Domain Experts from Large ML Models by Identifying Relevant Subspaces
}

\author[tub,eliza]{Pattarawat Chormai}
\author[tub,bifold]{Ali Hashemi}
\author[tub,bifold,korea,mpi]{Klaus-Robert M\"uller}
\author[bifold,charite]{Gr\'egoire Montavon\corref{cor1}}
\ead{gregoire.montavon@charite.de}

\cortext[cor1]{Corresponding author}

\address[tub]{Machine Learning Group, Technische Universit\"{a}t Berlin (TU Berlin), 10587 Berlin, Germany}
\address[eliza]{Konrad Zuse School of Excellence in Learning and Intelligent Systems (ELIZA), 64289 Darmstadt, Germany}
\address[bifold]{BIFOLD\,--\,Berlin Institute for the Foundations of Learning and Data, 10587 Berlin, Germany}
\address[korea]{Department of Artificial Intelligence, Korea University, Seoul 136-713, Korea}
\address[mpi]{Max Planck Institute for Informatics, 66123 Saarbr{\"u}cken, Germany}
\address[charite]{Institute for AI in Medicine, Charit\'e\,--\,Universit\"atsmedizin Berlin, 10115 Berlin, Germany}

\begin{abstract}
Knowledge distillation involves transferring the predictive capabilities of large, high-performing AI models (teachers) to smaller models (students) that can operate in environments with limited computing power. In this paper, we address the scenario in which only a few classes and their associated intermediate concepts are relevant to distill. This scenario is common in practice, yet few existing distillation methods explicitly focus on the relevant subtask. To address this gap, we introduce `SubDistill', a new distillation algorithm with improved numerical properties that only distills the relevant components of the teacher model at each layer. Experiments on CIFAR-100 and ImageNet with Convolutional and Transformer models demonstrate that SubDistill outperforms existing layer-wise distillation techniques on a representative set of subtasks. Our benchmark evaluations are complemented by Explainable AI analyses showing that our distilled student models more closely match the decision structure of the original teacher model.
\end{abstract}


\end{frontmatter}

\section{Introduction}

Large general-purpose pretrained models, such as ResNets \cite{DBLP:conf/cvpr/HeZRS16}, BERT \cite{DBLP:conf/naacl/DevlinCLT19}, GPT-2 \cite{radford2019language}, and others \cite{DBLP:journals/tmlr/OquabDMVSKFHMEA24,Muttenthaler2025}, have become popular starting points for designing predictive machine learning models across various applications. These models, often referred to as ``foundation models'' \cite{DBLP:journals/corr/Bommasani21}, are trained using large computational resources and embody very rich high-dimensional representations. However, these foundation models, besides their higher computational footprint, must rely increasingly on specialized systems (e.g.\ inference servers \cite{DBLP:conf/hpec/Li0GT24}) and may become a failure point or pose privacy risks. More compact models that can be stored and evaluated locally may thus prove more flexible and resilient, provided that they can be trained up to sufficiently high accuracy.

\medskip

The field of knowledge distillation \cite{DBLP:conf/kdd/BucilaCN06,DBLP:journals/corr/HintonVD15} addresses the challenge of transferring knowledge from a large model to a smaller, more efficient one. Various algorithms have been proposed for this purpose, typically based on a teacher-student learning scenario where the student model is much smaller and is trained to replicate the output neurons of the teacher model (e.g.,\ \cite{DBLP:journals/corr/HintonVD15, DBLP:journals/corr/abs-1910-01108-distilledBert, DBLP:conf/acl/SunYSLYZ20, DBLP:journals/access/Du22}). A common formulation of the knowledge distillation problem is layer-wise distillation, where the representations of the teacher and student are matched not only at the output but also at intermediate layers. The layer-wise distillation approach underlies many of the state-of-the-art distillation techniques (e.g.,\ \cite{Miles_2024_CVPR,DBLP:conf/emnlp/JiaoYSJCL0L20,DBLP:conf/icml/LiangZZHCZ23}). While these approaches have resulted in significant accuracy gains compared to output-focused distillation approaches, the layer-wise paradigm becomes challenged when the objective is not to distill the whole teacher model, but only part of it, for example, if the user is interested in a narrow subset of classes.

\medskip

\begin{figure*}
 \makebox[\textwidth][c]{\includegraphics[width=1.1\textwidth]{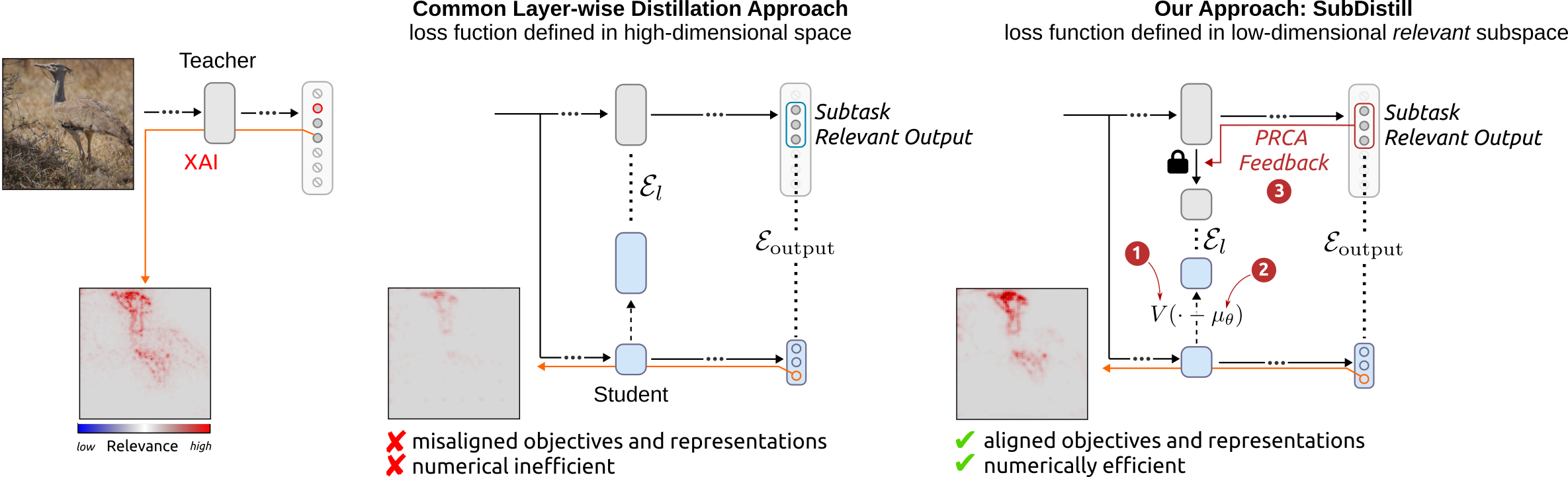}}
\caption{
Schematic depiction of our SubDistill approach tailored for subtask distillation. Our proposed method combines (1) orthogonal transformation, (2) centering, and (3) an Explainable AI analysis called PRCA \cite{chormai-tpami24} for identifying task-relevant components of the teacher's representation. Besides producing accurate student models, our approach also better preserves the teacher's decision strategy, as revealed by Explainable AI pixel-wise heatmaps.
}
\end{figure*}

Aiming to address this current gap in knowledge distillation, we propose `SubDistill', a layer-wise distillation method that identifies relevant components of the teacher's decision making prior to distillation. Our method defines a novel optimization objective and combines it with a recent Explainable AI method called `principal relevant component analysis (PRCA)' \cite{chormai-tpami24}, to enable an efficient transfer of relevant information from the teacher to the student. Compared to classical knowledge distillation approaches, our proposed SubDistill method adds guarantees on student-teacher representational alignment, improves numerical properties of the distillation algorithm, and aligns objectives at each layer of distillation to unambiguously focus on distilling features that are truly task-relevant.

\medskip

Across CIFAR-100 and ImageNet subtasks and diverse teacher-student pairs (CNNs and Vision Transformers \cite{DBLP:conf/iclr/DosovitskiyB0WZ21}), our `SubDistill' approach delivers consistent accuracy gains over existing distillation techniques across different models, distillation subtasks, and dataset sizes. Furthermore, our results are complemented with an Explainable AI analysis, which demonstrates that, beyond test set accuracy, our method also better preserves the decision structure of the teacher at the pixel level.
Our approach is defined at an abstract level and, in principle, applicable to a broad range of teacher and student architectures; its carefully crafted optimization objective combining Knowledge Distillation and Explainable AI makes our method robustly applicable without resorting to extensive hyperparameter searches.
Code for our paper is available at \href{https://github.com/p16i/subdistill}{github.com/p16i/subdistill}.

\section{Related Work}

In this related work section, we focus on the subfields of knowledge distillation (KD) that are most relevant to our work, particularly those concerning the use of representations from the teacher model and the intersection of KD and Explainable AI. For a broader overview of the field of KD, we refer the reader to recent surveys \cite{DBLP:journals/pami/WangY22,DBLP:journals/ijcv/GouYMT21, mansourian2025a}. 

\subsection{Knowledge Distillation Guided by Internal Structures}
\label{section:lkd}

Many studies concentrate on aligning the student's representation closely with that of the teacher. 
\cite{DBLP:journals/corr/RomeroBKCGB14} proposes the first method for guiding the student to replicate the teacher's intermediate representation, leveraging learnable linear adapters.
Subsequently, other works introduce more complex transfer objectives and adapter functions to enhance transfer efficacy. 
For instance, \cite{DBLP:conf/cvpr/YimJBK17} proposes to  conduct the transfer by examining the connections between the representations of successive layers, while  \cite{DBLP:conf/iccv/HeoKYPK019} derives a transfer module that prevents the student from absorbing uninformative features from the teacher.
\cite{DBLP:conf/cvpr/AhnHDLD19,DBLP:journals/tnn/PassalisTT21}  formulates transfer objectives using information-theoretical measures, whereas \cite{DBLP:journals/corr/HuangW17a} frames the transfer as a matching problem between the student’s and teacher’s distributions.
Other studies provide guidance through the  statistics or structures of representations.
Such works include  aligning teacher and student attention maps, which can be constructed from the norm of the representation  \cite{DBLP:conf/iclr/ZagoruykoK17} or the gradient statistics \cite{DBLP:conf/icml/SrinivasF18}.
In addition, some work proposes preserving the relational \cite{DBLP:conf/cvpr/ParkKLC19,DBLP:conf/iccv/TungM19} or topological \cite{kim2024do-topo-transfer} structures of the teacher and student representations; or aligning the Gram matrices of the teacher and student \cite{dasgupta2025improving}.
While these works aim to  improve student-teacher representation alignment, they do not specifically address the scenario where only a subset of the teacher is relevant to distill.

\medskip
 
More closely related to our specific aims in this paper is the work of \cite{DBLP:conf/nips/KimPK18,rdimkd}, which proposes to first extract factors from the teacher's representation and use them to guide the student.
In a similar vein, \cite{DBLP:conf/icml/LiangZZHCZ23} broadens the concept to derive task-specific insights from the teacher, while \cite{DBLP:conf/aaai/ZhouZ025} proposes a mechanism that guides the student towards concentrating on  principal components that are important for their task.
 \cite{DBLP:conf/aaai/MilesM24,Miles_2024_CVPR} investigate the influence of adapter functions and their properties on knowledge transfer. In comparison to these works, we leverage recent techniques from the field of Explainable AI to robustly and systematically extract the relevant subspace at each layer. Furthermore, our approach is embedded in an overall numerically efficient optimization objective, which our experiment finds to be crucial to converge to stable and well-generalizing student models.

\subsection{Knowledge Distillation and Explainable AI}

Recent works have aimed to make use of techniques developed in the field of Explainable AI (XAI) \cite{DBLP:journals/aim/GunningA19,DBLP:journals/pieee/SamekMLAM21,DBLP:journals/inffus/ArrietaRSBTBGGM20} to either better understand the distillation process or to improve it. One line of work focuses on investigating whether specific characteristics of the teacher model or its interpretative properties are preserved in the student. For example, \cite{DBLP:conf/nips/OjhaLRLL23} conducts  a comprehensive study on which specific properties are transferred during KD, including localization capabilities, adversarial vulnerability, and data variance, while  \cite{pmlr-v202-han23b} looks at the number of unique concept detectors developed by the student model.
\medskip

More directly related to our work is the encouragement for students to have the same  explanation as the teacher. In particular, \cite{Guo_2023_CVPR-cam-transfer} demonstrates that the principle allows the student to learn efficiently when the training data is limited, while \cite{DBLP:journals/corr/abs-2402-03119-e2-kd,bassi2024explanationneeddistillationmitigating} illustrates its benefit in ensuring that the student learns the right features.  
On the other hand, \cite{wu2023mechanisms} examines the extent to which the mechanisms of the teacher are transferred to the student during knowledge distillation and finds that distilling with Jacobian matching  \cite{DBLP:conf/icml/SrinivasF18} improves the degree of transfer. While our method leverages Explainable AI in the process of distillation, it specifically leverages the ability of recent Explainable AI techniques to extract representation subspaces that are specific to a particular subtask.

\medskip
Finally, similar to us, the method of
\cite{DBLP:journals/pr/YeomSLBWMS21} uses XAI to find and remove its irrelevant feature maps. However, it is a pruning technique, thus requiring student and teacher to have the same overall architecture, whereas our approach is a distillation method and thereby not subject to this restriction.

\section{Proposed Method for Subtask Distillation: `SubDistill'}
\label{section:method}
We address the question of how to distill a large teacher model $f_T$ into a student model $f_\theta$ when only specific tasks of the teacher are relevant to distill, say, the student only needs to cover a specialized subset of the teacher's classes. Our starting point is the well-established layer-wise knowledge distillation scheme  \cite{DBLP:journals/corr/RomeroBKCGB14,vaswani2017attention,DBLP:conf/cvpr/AhnHDLD19,mansourian2025a} that operates both on the outputs and the intermediate activations. The teacher and student are viewed as two models composed of $L$ layers, with the last layer being the output. The overall error to minimize is defined as:
\begin{align}
    \label{eq:model-distillation-loss}
    \mathcal{E}_\text{output}(a_T^{(L)}, a_\theta^{(L)})  +
        \sum_{l=1}^L \alpha_l \cdot \mathcal{E}_l(a_T^{(l)},a_\theta^{(l)})
\end{align}
where $\Err_\text{output}$ is the KL divergence between the teacher's and student's probability \cite{DBLP:journals/corr/HintonVD15}, and $\mathcal{E}_l$ quantifies the mismatch between the teacher activations $a_T^{(l)}$ and student activations $a_\theta^{(l)}$; the latter can be seen as a function of the student's parameters $\theta$. The coefficients $\alpha_l$'s are weighting terms for the loss functions at each layer and need to be tuned appropriately. Our proposed `SubDistill' method extends this basic layer-wise distillation scheme in two ways:
\begin{enumerate}[label=(\roman*)]
\setlength{\itemsep}{0pt}
    \item it formulates a novel `orthogonal subspace matching' loss that is designed for advantageous numerical properties and to ensure accurate teacher-student representational alignment;
    \item it steers the orthogonal subspaces towards task-relevant components, leveraging for this an recent Explainable AI technique called `principal relevant component analysis' (PRCA) \cite{chormai-tpami24}.
\end{enumerate}
These two building blocks form the core of SubDistill. Taken together, they aim to enable a robust transfer of information from the teacher (especially, its subtask knowledge) to the student, and they are presented in Sections \ref{section:orthogonal-subspace-matching} and \ref{section:prca} respectively.

\subsection{Orthogonal Subspace Matching}
\label{section:orthogonal-subspace-matching}
Consider a layer $l$, with $a_T$ and $a_\theta$ the teacher's and student's activation vectors at this layer of dimensions $d$ and $K$ respectively. Denote by $\mu_T$ and $\mu_\theta$ the corresponding mean activation vectors of same dimensions ($\mu_\theta$ can be estimated using the mini-batch mean \cite{schraudolph2002centering} due to the evolving nature of the student). We propose the following formulation for the loss at this layer:
\begin{align}
  \Err_l(a_T, a_\theta) = \mathbb{E}  \big [  \big \| 
     V(a_\theta - \mu_\theta) - U^\top (a_T - \mu_T)
    \big \|^2_2   \big ]
    \label{eq:ours}
    \end{align}
where $V$ is a matrix of parameters of size $K \times K$ and that we constrain throughout the distillation process to satisfy orthogonality ($V^\top V= I_K$). The expectation in Eq.\ \eqref{eq:ours} denotes an average over the training data and the various random factors entering into the student and teacher models' predictions. The matrix $U$ of size $d \times K$ consists of $K$ orthogonal columns and is used to define what in the teacher representation is relevant to be transferred to the student. (We propose a specific instantiation of the matrix $U$ in Section \ref{section:prca}.) We refer to the matrices $U$ and $V$ mapping the student to the teacher and vice versa as the `\textit{adapter}' as they enable a matching between the teacher and student representations of different dimensionalities.

In practice, the orthogonality constraint $V^\top V = I_K$ can be realized by optimizing on the Stiefel manifold \cite{edelman1998geometry,DBLP:conf/icml/CasadoM19} (which is the choice we opt for) or having the soft penalty $ \|V^\top V  - I_K \|_F^2$ \cite{bansalortho} added to Eq.\ \eqref{eq:ours}. Lastly, to ensure that the loss function  at each layer contributes equally to the overall error, we define $\alpha_l = \alpha \,/\, \E[ \|(U^{(l)})^\top (a_T^{(l)} - \mu_T^{(l)}) \|_2^2]$ for some $\alpha >0$. This leaves us with a single parameter $\alpha$ determining how the layer-wise losses are weighted compared to the output loss, thereby dramatically reducing the hyperparameter search space.

In the following, we analyze the properties of our proposed formulation in Eq.\ \eqref{eq:ours} and compare it to a common formulation that would consist of optimizing $\mathbb{E}[\big \|W a_\theta + b - a_T \|^2]$, referred to as $(W,b)$-formulation in the following, and that encompasses \cite{DBLP:journals/corr/RomeroBKCGB14,DBLP:conf/emnlp/JiaoYSJCL0L20,DBLP:conf/aaai/MilesM24,Miles_2024_CVPR}. In particular, the discussion of our method's advantages can be organized along three categories.

\paragraph{Representational Alignment} One can show that minimizing the layer-wise objective of Eq.\ \eqref{eq:ours} ensures alignment between the student's and teacher's representation. In particular, perfect reconstruction obtained through minimizing the objective in Eq.\ \eqref{eq:ours} implies a perfect representational alignment---more specifically, linear centered kernel alignment (CKA) \cite{DBLP:conf/icml/Kornblith0LH19}---between $a_\theta$ and $U^\top a_T$. Formally, $\Err_l(a_T, a_\theta) = 0$ implies $\mathrm{CKA}(\{U^\top a_T - \mu_T\},\{ a_\theta-\mu_\theta\}) = 1$. A proof is given in Supplementary Note A. It is important to note that the more basic $(W,b)$-formulation does not satisfy this property. It is the specific introduction of orthogonality in the adapter, similar to \cite{Miles_2024_CVPR}, that guarantees this property, and which ensures that the modeling occurs in the student rather than in the adapter. Hence, the orthogonality property is key in maintaining close correspondence between the student and the teacher.

\paragraph{Numerical Aspects} The optimization of our objective is significantly easier than the classic $(W,b)$-formulation. The centering term $\mu_\theta$ in Eq.\ \eqref{eq:ours} plays a key role in this respect, avoiding pathological curvature of the objective function. Mathematically, one can show that the principal directions of curvature for the $(W,b)$-formulation are given by the eigenvalues\,/\,eigenvectors of the matrix $\mathbb{E}[a_\theta a_\theta^\top]$ (cf.\ Supplementary Note B). When $a_\theta$ has non-zero mean, the gap between the top eigenvalue and lower eigenvalues tends to be large, giving the error function an elliptical shape that is difficult to optimize \cite{bishop1995neural}. The importance of centering is also emphasized in earlier works such as \cite{lecun2002efficient,schraudolph2002centering}. A further advantage of our approach compared to e.g.,\  \cite{Miles_2024_CVPR} is that optimization occurs on a lower-dimensional set of parameters. In particular, it is numerically more efficient to maintain the orthogonality constraint in this reduced dimensional subspace.

\paragraph{Objective Alignment} The additional degree of freedom given by our method in specifying the projection matrix $U$ helps address another limitation of the basic $(W,b)$-formulation and potentially any distillation method that is not tailored for subtask distillation and equipped with some form of top-down feedback. In such a formulation, the reconstruction objective requires reconstructing everything about the representation, irrespective of whether some aspects are task-irrelevant. This creates a mismatch between the objectives at each layer---which has to be arbitrated by a careful balancing of the loss functions at each layer---and thus potentially extensive hyperparameter searches. By an appropriate choice of matrix $U$ at each layer, our method largely mitigates these issues, enabling losses to align on the shared goal of distilling subtask-relevant features at each layer and reducing the sensitivity of distillation to the exact choice of coefficients $\alpha_l$s.

\subsection{Identifying Subtask-Relevant Subspaces}
\label{section:prca}

With Eq.\ \eqref{eq:ours}, we have proposed a formulation of layer-wise distillation, which operates on a subspace $U$ of the teacher's representation. We motivated our approach in (1) its ability to ensure student-teacher representational alignment, (2) numerical properties, and (3) enabling an alignment of objectives at each layer towards the same goal of reconstructing the teacher's task-relevant subspace. However, we have so far assumed that such focus on relevant components of the teacher was embodied in the projection matrix $U$ without addressing how this matrix is actually obtained. To resolve this, we propose to leverage a recently proposed Explainable AI method, Principal Relevant Component Analysis (PRCA) \cite{chormai-tpami24}, which consists of extracting a subspace that maximally expresses some target quantity at the output of the network. We contribute an adaptation of PRCA for knowledge distillation, together with some stability improvements.

Whereas the original PRCA method identifies a subspace $U$ that maximally expresses a given class logit, we propose to identify a subspace $U$ that maximally expresses the teacher's classification outcome. Specifically, we look for a subspace $U$ that maximizes the margin $\Delta$ between the actual predicted class $j^\star$ and the class it ranks second $j^\dagger$, something that can be quantified by the two classes' log-probability ratio:
\begin{align}
    \label{eq:delta}
     \Delta 
     = \log \big[p(j^\star|x) \big/ p(j^\dagger|x) \big].
\end{align}
Considering a given layer of the teacher with (centered) activation vector $\widetilde{a}_T = a_T - \mu_T$, we can build at this layer $\widehat{\Delta} = \langle \widetilde{a}_T, c_T\rangle$, a homogeneous surrogate of the quantity defined in Eq.\ \eqref{eq:delta}, where $c_T$ can be a simple local evaluation of the gradient $\partial \Delta/\partial \widetilde{a}_T$ or a more advanced response model. Inspired by the PRCA method, we then look for a subspace (a matrix $U$ of orthogonal vectors) that best expresses the quantity of interest $\Delta$, more specifically, the surrogate quantity $\widehat{\Delta}$. We define for that purpose the objective
\begin{align}
    \max_U ~ \Big\{\E\big[ \langle \widetilde{a}_T, c_T \rangle_U\big] + \beta^{-1} \cdot \E\big[ \langle \widetilde{a}_T, \widetilde{a}_T \rangle_U\big]+ \beta \cdot \E\big[ \langle c_T,c_T \rangle_U\big] \Big\}
    \quad
    \mathrm{s.t.} \quad 
    U^\top U= I_K
\label{eq:prca}
\end{align}
where $\langle \cdot,\cdot \rangle_U = \langle U^\top \cdot, U^\top \cdot \rangle$ denotes the Euclidean dot product in some subspace $U$.
The first term of this optimization objective is the quantity of interest, and
the second and third terms are stabilization terms that we added to the original PRCA formulation. The optimization problem can be solved analytically. (The analytical solution, its derivation of the solution, and further justification for the introduced stabilization terms can be found in Supplementary Note C) The hyperparameter $\beta > 0$ controls the extent by which the subspace $U$ is determined by the model response rather than the input activations. In the extreme case where $\beta$ is set close to zero, the optimization problem reduces to a PCA analysis of the activations, without considering any dependence on the model response. Empirically, we find that setting $\beta = \sqrt{\operatorname{tr}(\Sigma_a)/\operatorname{tr}(\Sigma_c)}$ works well in all tested distillation scenarios and has a simple geometric interpretation: it is equivalent to first dividing $\widetilde{a}_T$ and $c_T$ with the square root of their respective (uncentered) total variance
and then setting $\beta = 1$, thereby ensuring that neither the variance of activations nor the model response spuriously dominates the objective.

\subsection{Verification of the Method's Design on Synthetic Data}
\label{section:synthetic-data}

\begin{figure}[t!]
    \centering
    \includegraphics[width=\linewidth]{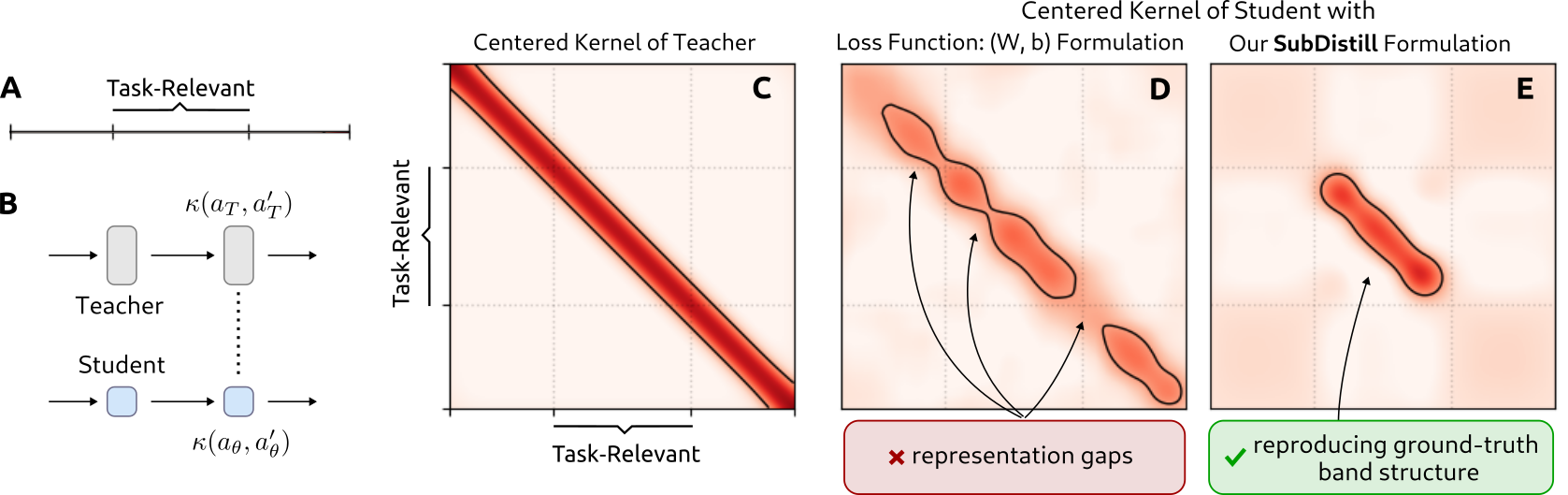}
    \caption{
    Demonstration of SubDistill on a synthetic example where the data follows a one-dimensional manifold, and where only a fraction of the (one-dimensional) input domain possesses task-relevant features (A). Distillation requires the student to reproduce the teacher's internal representation, which we subsume by its kernel (B). The teacher's kernel exhibits a perfect band structure, with only its middle part being task-relevant (C). The classical $(W,b)$-formulation fails to focus on the relevant section of the kernel and numerical issues further hinder distillation, leading to sporadic gaps in its band structure (D). Only the SubDistill approach, owing to its advantageous numerical properties and its focus on what is task-relevant, is able to reliably capture the teacher's band structure (E).
    }
    \label{fig:toy-band}
\end{figure}

To illustrate our method, its numerical stability, and its ability to align with the teacher's decision making, we construct a synthetic experiment where a student is tasked to adapt its representation to that of the teacher through a linear adapter function. The setup of the experiment is depicted in Fig.~\ref{fig:toy-band} (B). 

In this example, the data is constructed as a long one-dimensional manifold (A), which appears as a band structure in the teacher's associated kernel, shown in Fig.\ \ref{fig:toy-band} (C), i.e.,\ $\kappa(a_T, a'_T) = \widetilde{a}_T^\top \widetilde{a}_T'$. This band structure highlights that consecutive instances on the data manifold are rightly considered similar, and remote instances are rightly considered dissimilar. 
Additionally, as the marker in (A) indicates, only instances in the middle of the manifold are assumed to be relevant. 
In the ideal case, successful distillation should therefore produce a student representation with an associated kernel exhibiting similar band structure but only in the relevant area.
Fig.~\ref{fig:toy-band} (D) shows the behavior of the common $(W,b)$-formulation that defines the loss function in the high dimensional space of the teacher. 
We can see that the $(W,b)$-formulation student attempts to recover every part of the teacher's representation but fails, as seen from its kernel exhibiting an inhomogeneous band structure. In contrast, the centered kernel of the student distilled via our SubDistill method has a well-aligned band structure and concentrates in the right region, as can be seen in Fig.~\ref{fig:toy-band} (E).
We provide the details of the experiment in Supplementary Note D.

\section{Empirical Evaluation}
\label{sec:experiments}

We now proceed with evaluating our SubDistill method's capability in real distillation scenarios involving popular image datasets and competitive neural network models. To emphasize the focus of our study on subtask distillation, we consider the scenario where the student has to mimic the decision of a significantly larger teacher on a subset of classes this time much smaller than the original set of classes the teacher is trained on.

\paragraph{Subtasks} To construct such subtasks, we leverage  the class taxonomy  of  CIFAR-100 and ImageNet.
For both datasets, we generate three representative subtasks, where each subtask is generated by selecting one superclass, and defining as subtask the classification of instances within that superclass. For CIFAR-100, the superclasses are chosen according to accuracies achieved by a ResNet18 \cite{DBLP:conf/cvpr/HeZRS16}, namely `People' (hardest), `Insects', and `Vehicles\,2' (easiest). For ImageNet, we proceed similarly and, based on the accuracies of a   
ResNet18 from TorchVision \cite{torchvision2016}, we choose `Domestic\,Cat' (hardest), `Truck', and `Wading\,Bird' (easiest).

\paragraph{Distilled Models} We consider a selection of teacher/student pairs ranging from well-tested architectures commonly used in benchmarks to more realistic scenarios involving specialized architectures. On ImageNet, we consider the following architecture pairs:
\begin{enumerate}[label=(\roman*)]
\setlength{\itemsep}{0pt}
\item ResNet18 \cite{DBLP:conf/cvpr/HeZRS16} $\to$ ResNet18-S: In this setting, we distill a standard pretrained ResNet from TorchVision into a student model with same architecture but where the number of feature maps at each layer has been artificially limited to 32 in the first three blocks and to 24 in the last block.
\item WResNet101 \cite{DBLP:conf/bmvc/ZagoruykoK16} $\to$ MbNetv4 \cite{mobilenetv4}: In this setting, we distill a significantly larger and deeper model into MbNetv4, a specialized neural network architecture designed for mobile applications. This setting is more challenging because it prevents distillation techniques from assuming common internal structures between the teacher and student.
\item ViTB16 \cite{DBLP:conf/iclr/DosovitskiyB0WZ21} $\to$ EffFormerv2 \cite{li2022efficientformer}: This last setting resembles the previous setting, but Transformer architectures are this time used for the teacher and the student.
\end{enumerate}

\paragraph{Distillation Baselines} We benchmark our SubDistill method against classical output-based distillation (referred to as `Output Only') and a selection of  established or more recent layer-wise distillation techniques:
\begin{enumerate}[label=(\roman*)]
\setlength{\itemsep}{0pt}
    \item Attention Transfer (AT) \cite{DBLP:conf/iclr/ZagoruykoK17}, an established method that performs layer-wise alignments by guiding the student to mimic how the teacher attends to the input, serving as a prototypical technique of attention-based feature matching approaches;
    \item VID \cite{DBLP:conf/cvpr/AhnHDLD19}, a strong baseline that formulates the layer-wise loss through the maximization of the mutual information between the student and the teacher representations, reassembling methods that focus on statistical dependency rather than explicit feature alignment;
    \item VKD \cite{Miles_2024_CVPR}, a recent state-of-the-art technique that utilizes  orthogonal adapters and task-specific normalizations to encourage the student to preserve the representation structure of the teacher. This method exemplifies modern structure-preserving distillation techniques.
\end{enumerate}
In all our experiments and for each layer-wise distillation technique, we employ the same four distinct pairs of teacher–student layers to compute the layer-wise losses $\Err_l$ and use the same $\Err_\text{output}$ \cite{DBLP:journals/corr/HintonVD15}.
To keep the search over the hyperparameters $\alpha_l$ manageable, we fix all $\alpha_l = \alpha$ and select the optimal value from $\{10^{-2}, 10^{-1},..., 10^{2} \}$ (which contains the values used in \cite{DBLP:conf/cvpr/AhnHDLD19,DBLP:conf/iclr/ZagoruykoK17,Miles_2024_CVPR}) using a validation set. We provide the details of training and the correspondence between teacher and student layers  in Supplementary Note E and the estimation  of SubDistill's subspace $U$ in Supplementary Note F.

\subsection{Quantitative Results}
\label{section:imagenet-subtasks}

We first proceed with comparing the accuracy of the distilled student on the ImageNet dataset for each pair of teacher/student architectures, subtask, and distillation techniques. Results are shown in Table \ref{tab:main-table} (top). From the table, we see  that the student produced by SubDistill shows competitive and generally superior performance compared to all baselines, with large gains (between 5 and 10 percentage points) achieved in the WResNet101$\,\to\,$MbNetv4 setting. 
\begin{table*}[t!]
\setlength{\tabcolsep}{10pt}
    \centering
    \footnotesize
    \begin{tabular}{lcccccccccc}
    \toprule
    &  
    & 
    Output
    & AT & VID & VKD & SubDistill &
      \textit{largest}
    \\
         & \textit{Ref.} 
            &  Only  
            & \cite{DBLP:conf/iclr/ZagoruykoK17}
            &  \cite{DBLP:conf/cvpr/AhnHDLD19}
            &  \cite{Miles_2024_CVPR}
            & (\textbf{ours})
    & \textit{err.}
    \\
     \midrule
\multicolumn{6}{l}{\textit{Experiment with different subtasks (training size: 80\%)}}\\[2mm]
ResNet18 $\to$ ResNet18-S\\ 
       ~~~Wading Bird & \textit{98.0} & 90.8 & 91.3 & \underline{92.3} & 91.9 & \textbf{93.6} & $\pm 1.1$ \\
             ~~~Truck & \textit{88.0} & 77.9 & 79.2 & 79.3 & \underline{79.6} & \textbf{81.5} & $\pm 1.3$ \\
      ~~~Domestic Cat & \textit{72.8} & 60.8 & 62.7 & \textbf{65.9} & 64.0 & \underline{65.7} & $\pm 1.4$ \\[1mm]
WResNet101 $\to$ MbNetv4
        \\
       ~~~Wading Bird & \textit{98.4} & 89.1 & \underline{91.3} & 90.5 & 91.3 & \textbf{96.8} & $\pm 1.3$ \\
             ~~~Truck & \textit{93.6} & 76.7 & 78.0 & \underline{78.4} & 78.4 & \textbf{89.1} & $\pm 2.3$ \\
      ~~~Domestic Cat & \textit{74.8} & 60.3 & 60.7 & 62.0 & \underline{66.4} & \textbf{73.1} & $\pm 1.3$ \\[1mm]
ViTB16 $\to$ EffFormerv2
        \\
       ~~~Wading Bird & \textit{99.6} & 92.9 & 92.4 & 93.9 & \underline{95.6} & \textbf{96.8} & $\pm 0.6$ \\
             ~~~Truck & \textit{94.4} & 81.9 & 81.6 & 86.1 & \underline{87.6} & \textbf{90.8} & $\pm 0.7$ \\
      ~~~Domestic Cat & \textit{78.8} & 66.3 & 66.9 & 73.5 & \underline{75.9} & \textbf{76.4} & $\pm 1.3$ \\
\midrule
\multicolumn{6}{l}{\textit{Experiment with smaller training sizes (wading bird subtask)}}\\[2mm]
ResNet18 $\to$ ResNet18-S\\ 
       ~~~Training size: 50\% & \textit{98.0} & 88.9 & 90.1 & \textbf{91.2} & 90.0 & \underline{90.7} & $\pm 1.6$ \\
       ~~~Training size: 25\% & \textit{98.0} & 83.5 & 84.5 & \underline{88.8} & 87.9 & \textbf{89.7} & $\pm 1.1$ \\[1mm]
WResNet101 $\to$ MbNetv4 \\ 
       ~~~Training size: 50\% & \textit{98.4} & 84.9 & 83.9 & 85.5 & \underline{86.7} & \textbf{95.1} & $\pm 2.0$ \\
       ~~~Training size: 25\% & \textit{98.4} & 77.1 & 78.1 & 77.2 & \underline{80.8} & \textbf{91.2} & $\pm 1.8$ \\[1mm]
ViTB16 $\to$ EffFormersv2 \\
       ~~~Training size: 50\% & \textit{99.6} & 91.1 & 89.9 & \underline{93.3} & 93.1 & \textbf{95.9} & $\pm 0.9$ \\
       ~~~Training size: 25\% & \textit{99.6} & 84.9 & 84.8 & 86.8 & \underline{88.9} & \textbf{92.9} & $\pm 1.3$ \\
\midrule
\multicolumn{6}{l}{\textit{Experiment with a different dataset (C{\scriptsize{}IFAR}-100, training size: 80\%)}}\\[2mm]
ResNet18 $\to$ ResNet18-S 
\\
        ~~~Vehicles\,2 & \textit{96.2} & 90.2 & \underline{90.8} & 90.0 & \underline{90.8} & \textbf{94.7} & $\pm 0.7$ \\
           ~~~Insects & \textit{89.2} & 78.2 & 79.7 & 78.7 & \underline{82.5} & \textbf{85.3} & $\pm 1.1$ \\
            ~~~People & \textit{62.0} & 47.5 & 48.5 & 49.8 & \underline{52.9} & \textbf{60.9} & $\pm 1.9$ \\[1mm]
{WResNet40 $\to$ ResNet18-S}\\
        ~~~Vehicles\,2 & \textit{95.6} & 90.1 & 91.4 & \underline{91.7} & 91.3 & \textbf{93.7} & $\pm 0.7$ \\
           ~~~Insects & \textit{87.6} & 80.6 & 82.2 & 80.1 & \underline{82.7} & \textbf{84.1} & $\pm 1.3$ \\
            ~~~People & \textit{63.6} & 52.7 & 55.0 & \underline{55.5} & 55.1 & \textbf{59.4} & $\pm 1.3$ \\[1mm]
    \bottomrule
    \end{tabular}
    \\[0.2cm]
    \caption{
    Comparison between the accuracy of the student trained with our SubDistill method and other layer-wise distillation approaches on different subtasks and different teacher-student pairs.
    We report average accuracy from three random initializations. The column indicated with `Ref' is the subtask accuracy of the teacher, and the last column shows the largest standard error of each row.
    }
    \label{tab:main-table}
\end{table*}
Table \ref{tab:main-table} (middle) demonstrates that SubDistill continues to outperform alternatives when training with limited data. Our method produces the best student model in most configurations. While for conciseness we show results for the `wading bird' subtask, a comparable pattern is observed for the other two subtasks as well (See Supplementary Note G). Finally, Table \ref{tab:main-table} (bottom) extends the benchmark comparison to an additional dataset, CIFAR-100, and where we consider two additional teacher architectures: a ResNet18 model (which we adjust, similarly to \cite{chen2020simple}, to accommodate the smaller size of CIFAR-100 images and train ourselves) and a WideResNet40 provided by \cite{DBLP:conf/cvpr/KirchheimFO22}. Similar to the finding on the ImageNet subtasks, we see that the SubDistill performs generally substantially better than the baselines. The superiority of our method is particularly marked in the case of the ResNet18 pair on the People subtask, where it yields about 10 points improvement over the baselines.

Overall, the findings reported in this section clearly demonstrate the effectiveness and reliability of our SubDistill method for subtask distillation.
In the next section, we further examine various characteristics of SubDistill to gain deeper insights into the observed improvements.

\subsection{Analysis of the Student's Decision Strategy}
\label{sec:faithfulness-heatmaps}

Recent developments in learning large ML models (e.g.,\ foundation models \cite{DBLP:conf/fat/VaroquauxLW25,Dippel24RudolfV}) devote tremendous efforts and resources to ensure that the predictions of the models  are trustworthy and reliable \cite{lapuschkin-ncomm19,DBLP:conf/nips/Ouyang0JAWMZASR22,DBLP:journals/natmi/KauffmannDRSMM25, Muttenthaler2025}.
As a result, it is desirable (and perhaps expected) that the process of distillation properly transfers the teacher's decision strategies to the student.
In this section, we therefore take a closer look at how the student trained by our SubDistill approach and others makes predictions \cite{baehrens2010explain} and to what extent their strategies resemble those of the teacher. 

More specifically, using Explainable AI techniques \cite{DBLP:journals/pieee/SamekMLAM21}, we compute for each prediction of the student model an explanation, a pixel-wise heatmap, which depicts how relevant each input pixel is for the target class. The student explanation is then compared to that of the teacher.
We focus this analysis on the setup of the ResNet18 pair, with the student distilled on the Wading Bird subtask, and where we consider the 80\% and 25\% training data regimes to investigate how the amount of training data influences the student's learned decision strategies. We use the LRP \cite{bach-plos15,DBLP:series/lncs/MontavonBLSM19} technique, which is well-established for generating pixel-wise explanations, and we follow LRP's common hyperparameter selection heuristics, using the LRP-$\gamma$ rule for convolution layers and the LRP-$\epsilon$ rule for fully-connected layers \cite{DBLP:series/lncs/MontavonBLSM19}.

\begin{figure}[t!]
    \centering
    \includegraphics[width=\textwidth]{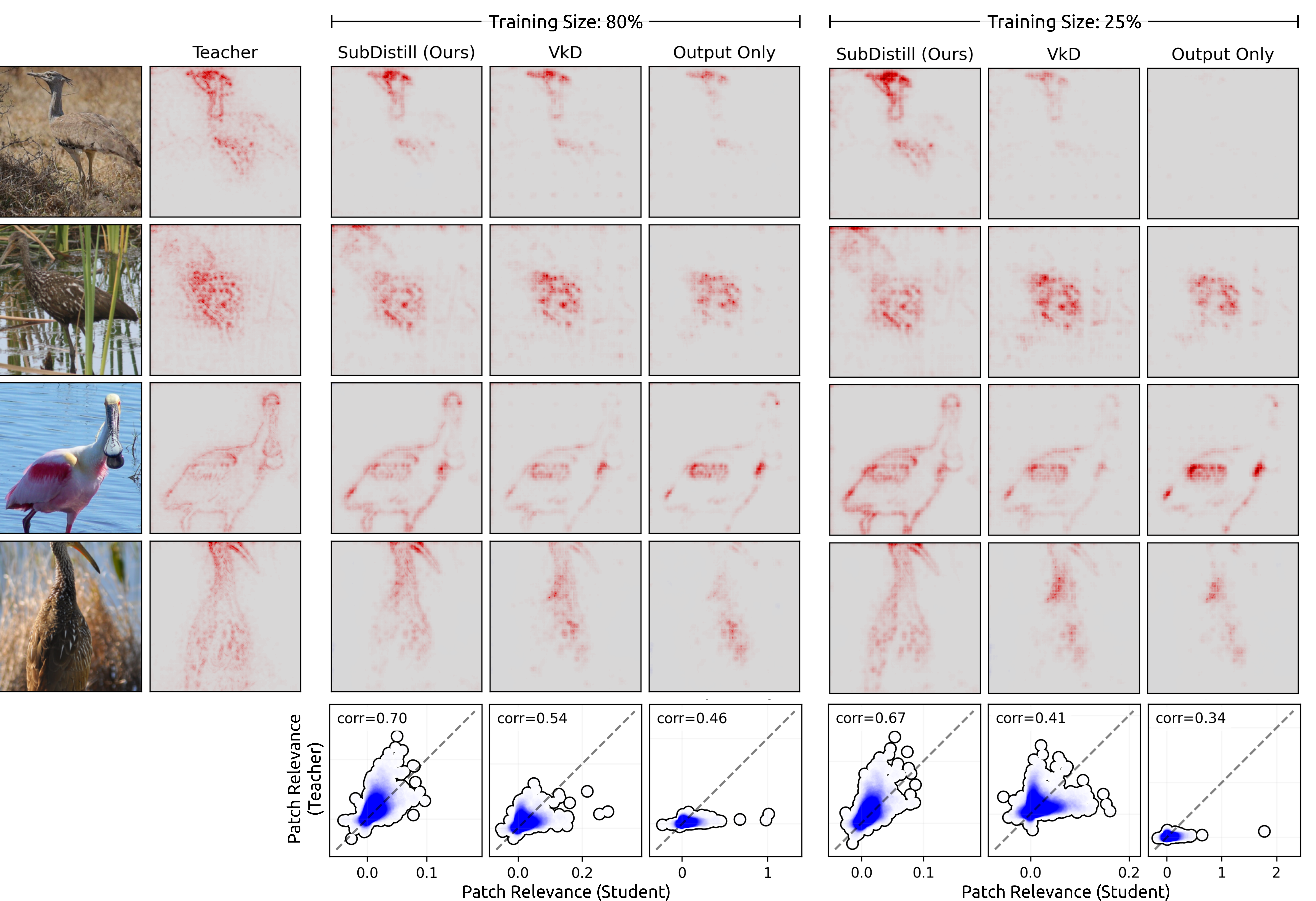}
    \caption{
    Top: Pixel-wise explanations of the predictions of the teacher and the distilled students trained on the ImageNet `wading bird' subtask using  80\% and 25\% of training data. Results are shown for a random selection of input images. We compare students produced by our SubDistill approach and the VKD, and `output only' baselines.
    Pixel-wise explanations are averaged over three training runs. 
    Bottom: Scatter plots comparing the teacher and the students explanations on a patch level (patches of size $8\times 8$) and for the whole data distribution. The diagonal line in the plots represents exact matching between teacher and student. Next to the scatter plots, we show the Pearson correlation coefficient (corr) between teacher and student explanations. 
    }
       \label{fig:heatmap-cossim}    
\end{figure}

Fig.~\ref{fig:heatmap-cossim} (top) shows the LRP attribution maps of the teacher and the student trained  by our SubDistill approach, VKD (which has the highest prediction accuracy among the baselines), and Output Only.
Across the two training data regimes, we see qualitatively that the attribution maps for the student learned by SubDistill tend to be more  similar to the ones of the teacher than those of the other students. This can be observed clearly in the third and fourth images, where only SubDistill is capable of maintaining the use of the head and beak features in the distilled decision strategy.
These varying levels of alignment between student and teacher across methods can be verified quantitatively by generating scatter plots, shown in Fig.~\ref{fig:heatmap-cossim} (bottom), where each point indicates  a given image patch in which its x- and y-coordinates are determined by its importance according to the student and teacher, respectively. We can observe that the SubDistill student correlates well with the teacher, performing much better than the other students. Furthermore, we see from this quantitative analysis that the gap between SubDistill and baseline distillation methods widens when using only 25\% of the data for training, showing that our method remains especially robust in the low data regime.
Whereas we have focused here on the Wading Bird subtask, we find the same pattern on the remaining two ImageNet subtasks, as well as when computing attribution maps with alternative approaches \cite{DBLP:conf/icml/SundararajanTY17,DBLP:journals/jmlr/StrumbeljK10} (See Supplementary Note G).
Overall, the results of this analysis highlight a unique aspect of our SubDistill approach, in which it does not only transfer the high predictive performance of the teacher but also does it in such a way that the student's  decision strategies deviate less from those of the teacher.

\section{Ablation Experiments}
\label{sec:ablations}

We aim to tease apart the contribution of each element in our SubDistill framework in this section.
These elements include  centering, squared norm normalization for $\alpha_l$, the choice of the projection matrix $U$, whether the transformation to a low-dimensional space is necessary (i.e., $\Err_l$ is defined in high dimension).
To isolate each element's contribution and keep computational requirements feasible, we perform the study by eliminating or changing  each element from the proposed formulation of Eq.\ \eqref{eq:ours} one at a time. For the first two elements, we set $\mu_\theta = 0$ and the normalization constant equal to $1$ respectively. For the subspace choice, we substitute the subspace $U$ with that of PCA or random orthogonal vectors. For the necessity of the low-dimensional space, we set $U=I_d$ and $V = W$ with $W^\top W = I_K$  using either  the soft penalty   or optimization on the Stiefel manifold.

Table \ref{tab:ablation} shows the results of these ablations. We see from the table that discarding centering or normalization leads to severe performance degradation, confirming our design choices. When discarding dimensionality reduction but maintaining orthogonality, both tested alternatives (v1: adding an orthogonality-inducing soft-penalty $1000 \cdot \|W^\top W - I_K\|_2^2$ and v2: maintaining orthogonality via high-dimensional Stiefel projection) lead to lower accuracy. The alternative v2 fares better in terms of accuracy but also has a much higher computational cost. 

Finally, when discarding the PRCA-based subspace extraction, the outcome depends on the subspace selection alternative. If we replace PRCA by a random orthogonal subspace (v1), we observe again a substantial loss of accuracy. When instead replacing PRCA by classical PCA (v2), the performance decrease varies across different classes, with one observed case of increase. As PCA can be seen as a special case of PRCA (with PRCA's parameter $\beta \approx 0$), this result suggests that a subtask-specific adaptation of PRCA hyperparameters, informed by how the subtask differs from the other tasks, might deliver the most robust results in practice.

\begin{table}[t!]
 \footnotesize
    \centering
    \makebox[\textwidth][c]{
    \begin{tabular}{lcccccccccc}
    \toprule
          & \thead{SubDistill\\(\textbf{ours})}
            & \thead{without\\centering} & \thead{without\\norm.}
            & \thead{without\\DimRed\\(v1)} 
            & \thead{without\\DimRed\\(v2)} 
            & \thead{without\\PRCA\\(v1)} 
            & \thead{without\\PRCA\\(v2)}
            & \thead{largest\\err.}\\
    \midrule
\multicolumn{3}{l}{CIFAR100 / ResNet18 $\to$ ResNet18-S}
\\
        ~~~Vehicles 2 & \textbf{94.7} & 88.8 & 93.6 & 93.7 & 94.5 & \underline{94.6} & 94.3 & $\pm 0.5$ \\
           ~~~Insects  & \textbf{85.3} & 77.8 & 83.7 & 82.2 & 84.5 & 84.2 & \underline{84.9} & $\pm 0.7$ \\
            ~~~People  & \textbf{60.9} & 47.7 & 59.2 & 58.5 & 59.0 & 57.7 & \underline{59.3} & $\pm 1.9$ \\[1mm]
\multicolumn{3}{l}{ImageNet / WResNet101 $\to$ MbNetv4}
\\
       ~~~Wading Bird &  \textbf{96.8} & 93.3 & 94.7 & 94.4 & 95.6 & 96.0 & \underline{96.5} & $\pm 0.5$ \\
             ~~~Truck &  \textbf{89.1} & 83.6 & 84.5 & 85.9 & 87.3 & 86.3 & \underline{88.9} & $\pm 1.5$ \\
      ~~~Domestic Cat &  73.1 & 69.7 & 71.9 & 70.0 & \underline{73.2} & 71.7 & \textbf{75.1} & $\pm 1.3$ \\
    \bottomrule
    \end{tabular}
    }
    \caption{
    Ablation study of our SubDistill in which its various components are discarded one at a time. The first column indicates the performance of our original method. The second column, 
    `without centering', refers to setting $\mu_\theta=0$ in Eq.\ \eqref{eq:ours}. `Without normalization' removes data-dependent scaling factor from the definition of $\alpha_l$.
    The next two columns `DimRed' remove the dimensionality reduction step and resort to the soft penalty (v1) high-dimensional projections (v2) to maintain orthogonality.  Lastly, we experiment with removing the specific PRCA projection, replacing it either with a random prediction (v1) or by PCA (v2).
    }
    \label{tab:ablation}
\end{table}

\subsection*{Choice of Layers for Guidance}
As a further ablation experiment, we would like to test the influence on the performance of the student of removing specific layer-wise losses $\Err_l$ from the overall distillation objective. To this end, we consider four different scenarios in which we apply the layer-wise loss (of SubDistill or the baseline approaches) at only Layer $\{1\}$, $\{1,2\}$, $\{1,2, 3\}$, or $\{1, 2, 3, 4\}$ on the three CIFAR-100 subtasks (similar to the scenario from Table \ref{tab:main-table}).

\begin{figure}[t!]
    \centering
    \includegraphics[width=0.95\linewidth,clip,trim=0 0 0 1cm]{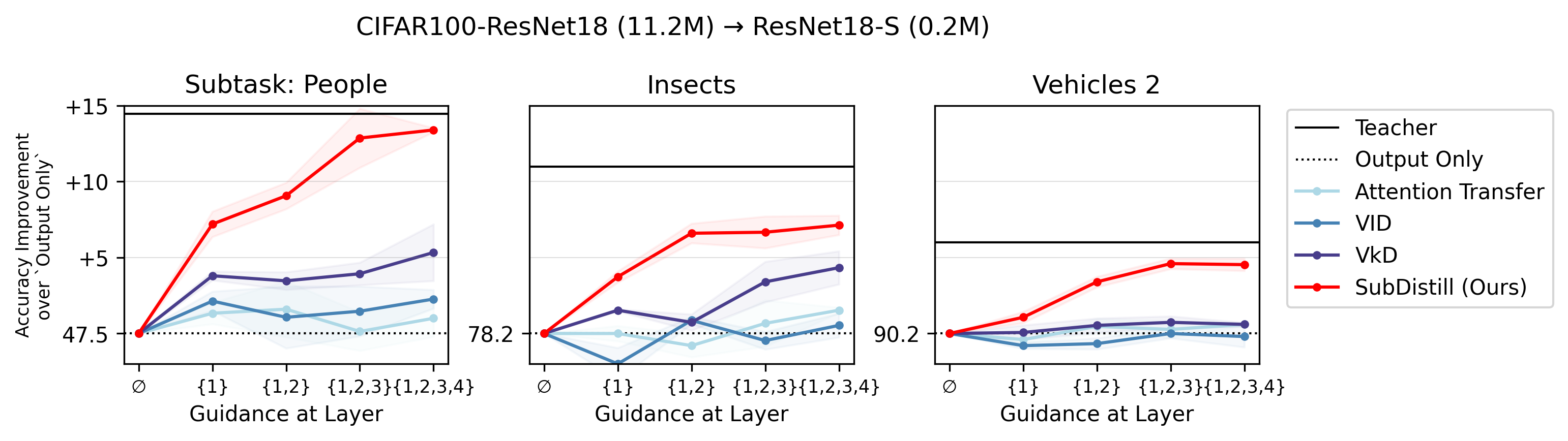}
    \caption{
    Ablation study on the layer-wise losses included in the overall distillation objective in Eq.\ \eqref{eq:model-distillation-loss}.
    The teacher is a ResNet18 trained on CIFAR-100 and the student is the ResNet18-S. The shaded region indicates standard error from three random initializations. From right to left, we remove the top-layer loss, the second-to-top layer loss, and so on until the input layer, at which point the distillation algorithm becomes equivalent to `output only'.
    }
    \label{fig:effect-layerwise-loss-locations}
\end{figure}

Fig.\ \ref{fig:effect-layerwise-loss-locations} shows the accuracy of the student trained with SubDistill and other baselines relative to that of only output supervision (denoted as $\varnothing$ here).
We observe that the SubDistill student shows progressive improvement as we incorporate guidance from more layers, and the incorporation of the early two layers tends to show  substantial improvement over output supervision.
 When guiding additionally with  later layers (e.g.,  $\{1,2,3\}$ and $\{1,2,3, 4\}$), although the improvement then becomes a plateau in some cases, it continues  for the People subtask. The observed trend here may be explained by the fact that the subtask is the most difficult one (from the perspective of the teacher), and the student thus needs  stronger supervision to learn well.
In contrast, the baseline methods show only marginal improvement and  demonstrate inconsistent performance trends. 
Overall, this ablation study  provides additional confirmation of SubDistill’s effectiveness at utilizing the teacher's representations  and offers insights into how the amount of representation supervision interacts with the difficulty of each subtask.

\section{Outlook: Subtask Distillation with Decoupled Training}
\label{section:decoupled-training}
Thus far, we have performed layer-wise knowledge distillation end-to-end.
Because the overall  learning objective consists of several loss terms, and determining the right balance between these terms (i.e.\ tuning the value of each $\alpha_l$) is a crucial step towards achieving good distillation results. In large scale settings (e.g., when the size of the  teacher is massive), heuristics for choosing $\alpha_l$ may show limits, and using as an alternative a full hyperparameter search over the different layers would cause compute requirements to grow exponentially with the number of layers. These limitations tend to cause bottlenecks in experimentation, slowing down overall model development cycles.

However, our SubDistill approach has a unique technical characteristic that can potentially evade the  need for exponential searches or subtle hyperparameter heuristics altogether. Namely, by decoupling the step of identifying what is relevant in the teacher's representations from the actual distillation step, SubDistill lends itself naturally to an alternative training procedure (which we call decoupled training) where the distilled model can be built in a layer-wise fashion
in a way that is reminiscent of earlier proposals for deep learning \cite{DBLP:journals/neco/HintonOT06,bengio2006greedy}. 
As a proof of concept, we apply this training strategy to SubDistill on the CIFAR100  Vehicles 2 subtask and when distilling from the ResNet18 teacher to a smaller ResNet18 student. We compare the approach to the original SubDistill, as well as on a set of the same baselines as in Section \ref{sec:experiments}.
For each distillation procedure, we use the output loss $\Err_\text{output}$ and the layer loss $\Err_l$ at layers 1 and 3. 

\begin{figure}
    \centering
    \begin{minipage}[c]{0.44\textwidth}
        \!\!\!\includegraphics[width=\linewidth,clip,trim=0 0 1.5cm 1.15cm]{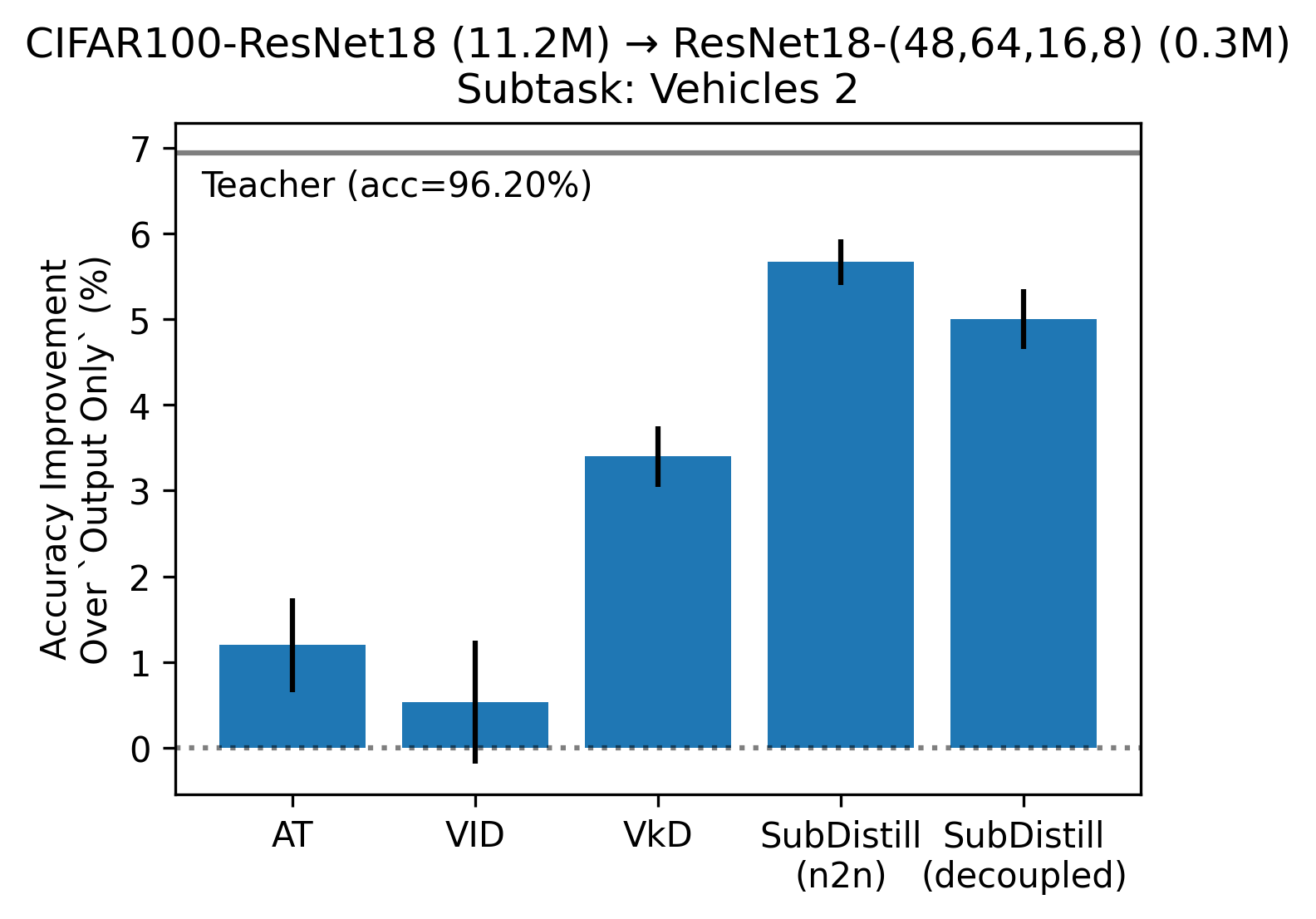}
    \end{minipage}~~~~
    \begin{minipage}[c]{0.44\textwidth}
    \caption{
    Comparison between the performance of students trained with SubDistill using decoupled  training and other layer-wise distillation approaches trained using  end-to-end training. Results are shown for distillation on the CIFAR100 Vehicles 2 subtask.
    The teacher model is ResNet18 (used in the main experiment), and the student is a small ResNet18 with layer dimensions equal to (48, 64, 16, 8). Accuracy scores are shown relative to the simple output-only baseline.
    }
    \vspace{1cm}
    \end{minipage}
    \label{fig:decoupled-distillation}
\end{figure}

Fig.\ \ref{fig:decoupled-distillation} shows the results. We see that not only does the SubDistill student with decoupled training require no tuning of $\alpha_l$'s, its performance remains competitive with SubDistill with joint training. More interestingly, it performs substantially better than the non-SubDistill baselines, although the latter benefit from joint training.
Overall, this result highlights that with a proper subtask-aware formulation of the distillation objective, the complex task of weighting the different layer-wise losses in the objective may become unnecessary. This provides immediate benefits in terms of runtime costs, but would also considerably foster reproducibility in the evaluation and benchmarking of distillation methods.

\section{Conclusion and Discussion}
Large, general-purpose ML models are becoming a commodity in many areas of machine learning. Yet, lightweight specialized models tailored for specific tasks remain desirable in a wide range of applications. While these domain experts could be in principle learned from scratch, distilling these `domain experts' general purpose models, brings multiple potential benefits, such as feature reuse and reduced exposure to dataset noise.

In this paper, we have proposed a novel distillation method, called SubDistill, which addresses a commonly overlooked aspect of knowledge distillation: the fact that only a limited portion of the large general-purpose model's behavior may be relevant to distill. Specifically, we leverage Explainable AI methods to carve low-dimensional task-relevant subspaces at each layer, on which the distillation process can more easily take place. Our method enables a better representational alignment between teacher and student, exhibits better numerical properties, and deconflicts the loss functions at each layer towards the same distillation objective.

Through an extensive set of experiments on multiple datasets and neural network architectures, we show that our SubDistill method systematically yields highly accurate models and demonstrates better robustness against the size of the training data. An Explainable AI analysis of the distilled models further reveals that SubDistill preserves the teacher's decision strategy, both qualitatively and quantitatively, better than other layer-wise knowledge distillation approaches.

Although SubDistill is designed abstractly enough to operate on any deep model and data modality, our investigation has so far been limited to image classification tasks. We see an adaptation to the text domain (e.g.,\ distilling lightweight low-vocabulary experts from LLMs) as a natural future work. Additionally, SubDistill could find applications in the medical domain: for example, large medical foundation models (e.g.,\ \cite{Dippel24RudolfV,chen2023meditron70b}) may be distilled into lightweight domain experts fitting on simple laptops, and capable of performing specific medical analyses, without relying on HPC and corresponding network infrastructure.

\section*{Acknowledgement}
This work was in part supported by the German Federal Ministry of Research, Technology and Space (BMFTR) under Grants BIFOLD24B, BIFOLD25B, 01IS18037A, 01IS18025A, and 01IS24087C.  P.C.\ is supported by the Konrad Zuse School of Excellence in Learning and Intelligent Systems (ELIZA) through the DAAD programme Konrad Zuse Schools of Excellence in Artificial Intelligence, sponsored by the Federal Ministry of Education and Research. K.R.M.\ was partly supported by the Institute of Information \& Communications Technology Planning \& Evaluation (IITP) grants funded by the Korea government (MSIT) (No. 2019-0-00079, Artificial Intelligence Graduate School Program, Korea University and No. 2022-0-00984, Development of Artificial Intelligence Technology for Personalized Plug-and-Play Explanation and Verification of Explanation).
We thank Jonas Lederer, Florian Schulz, and Manuel Welte for their useful feedback on the manuscript.

\bibliography{references}



\section*{Supplementary material}

# Distilling Lightweight Domain Experts from Large ML Models by Identifying Relevant Subspaces
## (Supplementary Notes)

Pattarawat Chormai, Ali Hashemi, Klaus-Robert Müller, Grégoire Montavon

**Supplementary Note A. SubDistill's Representation Alignment Guarantees**

We provide here formal results on the representation alignment that we discuss in the main text (Section 3). These results include the alignment guarantee of SubDistill (Proposition 1) and the necessity of orthogonality in the $(W, b)$-formulation (Proposition 2).

We first recall the notation used in the main text. The vector $a_T \in \mathbb{R}^d$ is the teacher's activation of some data point, and the vector $a_\theta \in \mathbb{R}^K$ is that of the student. The means of these two sets of vectors are $\mu_T$ and $\mu_\theta$ respectively. In the following, we write $A_T \in \mathbb{R}^{n \times d}$ and $A_\theta \in \mathbb{R}^{n \times K}$ to be the activations of the teacher and the student stacking from $n$ data points, and similarly write $\widetilde{A}_T$ and $\widetilde{A}_\theta$ to be their centered counterparts. We denote $\mathcal{E}_l(A_T, A_\theta)$ to be the average of applying the loss row-wise.

**Proposition 1** (Zero Loss Implies Perfect Alignment)**.** *Let $U \in \mathbb{R}^{d \times K}$ be a matrix with $K$ orthogonal vectors. When the loss function $\mathcal{E}_l(A_T, A_\theta) = 0$ (Eq. (2) in the main text), the centered kernels of the teacher's $\widetilde{A}_T U$ and the student's $\widetilde{A}_\theta$ are the same, and the linear centered kernel alignment [1] between them is maximized, i.e.,*

$$CKA(\widetilde{A}_T U, \widetilde{A}_\theta) = 1.$$

*Proof.* When $\mathcal{E}_l(A_T, A_\theta) = 0$, there exists $V \in \mathbb{R}^{K \times K}$ with $V^\top V = I_K$ s.t.

$$\widetilde{A}_T U = \widetilde{A}_\theta V^\top.$$

The centered linear kernel of the teacher is therefore

$$\widetilde{A}_T U (\widetilde{A}_T U)^\top = \left(\widetilde{A}_\theta V^\top\right) \left(\widetilde{A}_\theta V^\top\right)^\top = \widetilde{A}_\theta \widetilde{A}_\theta^\top.$$

By the definition of the linear CKA [1],

$$CKA(\widetilde{A}_T U, \widetilde{A}_\theta) = \frac{\|\widetilde{A}_\theta^\top \widetilde{A}_T U\|_F^2}{\|\widetilde{A}_T U (\widetilde{A}_T U)^\top\|_F \|\widetilde{A}_\theta \widetilde{A}_\theta^\top\|_F}$$

$$= \frac{\|\widetilde{A}_\theta^\top \widetilde{A}_\theta V^\top\|_F^2}{\|\widetilde{A}_\theta \widetilde{A}_\theta^\top\|_F \|\widetilde{A}_\theta \widetilde{A}_\theta^\top\|_F}$$

$$= 1,$$

where the last step uses the fact that the Frobenius norm is invariant to orthogonal transformation. □

---

**Assumption** (Optimal Bias of $(W, b)$-Formulation).

$$b = \mu_T - W\mu_\theta$$

The quantity above can be justified by minimizing $\mathcal{E}_l(A_T, A_\theta)$ with respect to $b$.

**Proposition 2** (Orthogonality is Necessary for $(W, b)$-Formulation). *With the assumption above and $\mathcal{E}_l(A_T, A_\theta) = 0$, attaining the maximum $CKA(\widetilde{A}_T, \widetilde{A}_\theta) = 1$ requires $W^\top W = \tau I_K$.*

*Proof.* Rewriting the $(W, b)$-formulation with the optimal bias leads to

$$\|Wa_\theta + b - a_T\|_2^2 = \|W(a_\theta - \mu_\theta) - (a_T - \mu_T)\|_2^2$$

When $\mathcal{E}_l(A_T, A_\theta) = 0$, there exists $W \in \mathbb{R}^{d \times K}$ s.t. $\widetilde{A}_T = \widetilde{A}_\theta W^\top$. The linear centered kernel of the teacher is

$$\widetilde{A}_T \widetilde{A}_T^\top = \widetilde{A}_\theta W^\top W \widetilde{A}_\theta^\top = \widetilde{A}_\theta \left( \sum_{k=1}^{K} \lambda_k u_k u_k^\top \right) \widetilde{A}_\theta^\top,$$

where $(\lambda_k, u_k)$'s are the eigenpairs of $W^\top W$. Because the linear CKA is invariant to isotropic scaling [1], the maximum is attained when $\widetilde{A}_T \widetilde{A}_T^\top = \tau \widetilde{A}_\theta \widetilde{A}_\theta^\top$ for some $\tau > 0$. Because $\sum_k u_k u_k^\top = I_K$, it therefore requires that all $\lambda_k = \tau$. □

### Supplementary Note B. SubDistill's Numerical Advantages

In this section, we further discuss additional numerical benefits of SubDitsill. We first show the direct connection between the eigenvalues and eigenvectors of the student's activation covariance $\Sigma_\theta$ and those of the Hessian matrix of $W$. We then provide a runtime comparison between our formulation and the $(W, b)$-formulation with orthogonality (which is necessary to get good alignment).

**Proposition 3** (Eigenvalues of Hessian). *Let $\Sigma_\theta = \mathbb{E}[a_\theta a_\theta^\top]$. The Hessian $H$ of the $(W, b)$-formulation of the loss function*

$$\mathcal{E}(a_T, a_\theta) = \mathbb{E}\left[\|Wa_\theta + b - a_T\|^2\right].$$

*with respect to $W$ is $H = 2(I_d \otimes \Sigma_\theta)$. The eigenvalues of $H$ are proportional to the ones of $\Sigma_\theta$ whose algebraic multiplicity is equal to $d$, and its eigenvectors can be written as Kronecker products of an arbitrary basis of the identity matrix with the eigenvectors of $\Sigma_\theta$.*

*Proof.* We define the Hessian to be

$$H = \partial^2_{\text{rvec}(W)} \mathcal{E}(a_T, a_\theta),$$

where rvec($\cdot$) denotes the row-major vectorization[1]. By expanding the squared norm and letting $\mathcal{C} = \mathcal{C}_0 + \mathcal{C}_1$ be all the zero- and first-order terms of $W$, the loss function becomes

$$\begin{aligned} \mathcal{E}(a_T, a_\theta) &= \mathbb{E}[a_\theta^\top W^\top W a_\theta] + \mathcal{C} \\ &= \mathbb{E}[\text{tr}(a_\theta^\top W^\top W a_\theta)] + \mathcal{C} \\ &= \text{tr}(W \Sigma_\theta W^\top) + \mathcal{C} \\ &= \sum_{i=1}^{d} W_{i,:}^\top \Sigma_\theta W_{i,:} + \mathcal{C}, \end{aligned}$$

where the third step uses the cyclic property of trace and the trace-expectation identity, and the vector $W_{i,:} \in \mathbb{R}^k$ denotes the $i$th row of $W$. Taking the derivative of the loss function w.r.t $W_{i,:}$ yields

$$\partial_{W_{i,:}} \mathcal{E} = 2\Sigma_\theta W_{i,:} + \mathcal{C}_1,$$

$$\implies \partial_{W_{i,:}} \partial_{W_{j,:}} \mathcal{E} = \begin{cases} 2\Sigma_\theta, & \text{if } i = j \\ 0, & \text{otherwise} \end{cases}.$$

Using the definition, the Hessian can therefore be written as a $dk \times dk$ block diagonal matrix

$$H = \begin{bmatrix} \partial_{W_{1,:}} \partial_{W_{1,:}} \mathcal{E} & 0 & 0 \\ 0 & \ddots & 0 \\ 0 & 0 & \partial_{W_{d,:}} \partial_{W_{d,:}} \mathcal{E} \end{bmatrix} = \begin{bmatrix} 2\Sigma_\theta & 0 & 0 \\ 0 & \ddots & 0 \\ 0 & 0 & 2\Sigma_\theta \end{bmatrix} = 2(I_d \otimes \Sigma_\theta),$$

concluding the proof of the first part.

The second part then follows immediately from a result in [2] (Theorem 4.2.12). The theorem shows that, for any square matrices $A$ and $B$ with eigenpairs $(x, \mu)$ and eigenpairs $(y, \nu)$ respectively,

$$\text{eigvals}(A \otimes B) = \{\mu\nu\}, \qquad \text{eigvecs}(A \otimes B) = \{x \otimes y\}.$$

$\square$

*Runtime Comparison.* We now present a runtime comparison between our SubDistill loss (Eq. (2) in the main text) and the $(W, b)$-formulation as well as its variants that enforce orthogonality on $W$ (i.e, $W^\top W = I_K$) either via the penalty $1000 \cdot \|W^\top W - I_K\|_2^2$ [3] (Soft) or optimizing on the Stiefel manifold [4, 5] (Hard). To this end, we conduct the benchmark on synthetic data in which we sample the student's $a_\theta$ from the $K$-dimensional standard normal distribution and the teacher's $a_T$ from the uniform distribution over $[-1, 1]^d$. For our loss, we construct $U$ from $K$ random orthogonal vectors. To collect runtime statistics, we perform one SGD update with batch size 128 and repeat the update for 70 trials (each with a different initialization of the loss's parameters); we consider the first 20 trials to be a warmup period and only use the statistics of the other 50 trials.

Fig. B.1 shows that our loss formulation shows substantially smaller runtime overhead than the hard variant of the $(W, b)$-formulation across scenarios. Although the soft orthogonal variant introduces the lowest overhead, it does not guarantee representation alignment, and the extent to which the property is satisfied potentially depends on the right strength of the penalty, which may require an extensive hyperparameter to adjust it properly.

---

[1] Let $X \in \mathbb{R}^{m \times n}$ and $X_{ij}$ be the entry in the $i$th row and $j$th column. We have rvec$(X) = [X_{11}, X_{21}, ..., X_{mn}]$.

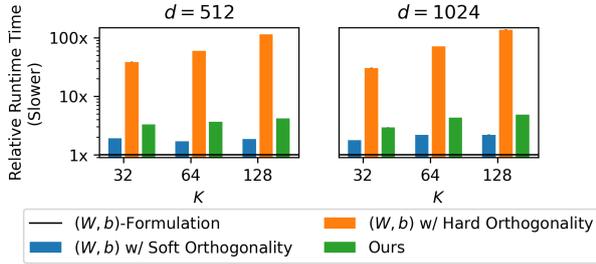

Figure B.1: Runtime comparison between our SubDistill loss (Eq. (2) in the main text) and different variants of the $(W, b)$ formulation. The statistics are collected from NVIDIA A100 (40GB).

## Supplementary Note C. Properties of Our Extended PRCA

We propose an analysis in Section 3.2 of the main text that extends the principle of PRCA [6] to find a subspace that preserves between-class margins, and we achieve the goal by regularizing the objective such that it is non-negative. In this supplementary note, we first discuss the importance of non-negativity and prove that our analysis satisfies it (Proposition 4). We then derive the analytical solution of the objective (Proposition 5).

To simplify the notation in this note, we drop the subscription and write the teacher's activation vector by $a$ (and also its centered version $\widetilde{a}$) and its associated response vector by $c$.

*Importance of Non-Negativity.* To understand the relationship between subspace dimensionality and subtask performance, we project activations on the subspace with various dimensionalities and track, in addition to the model's subtask performance, the margin (Eq. (3) in the main text) of each class when varying the dimensionalities.

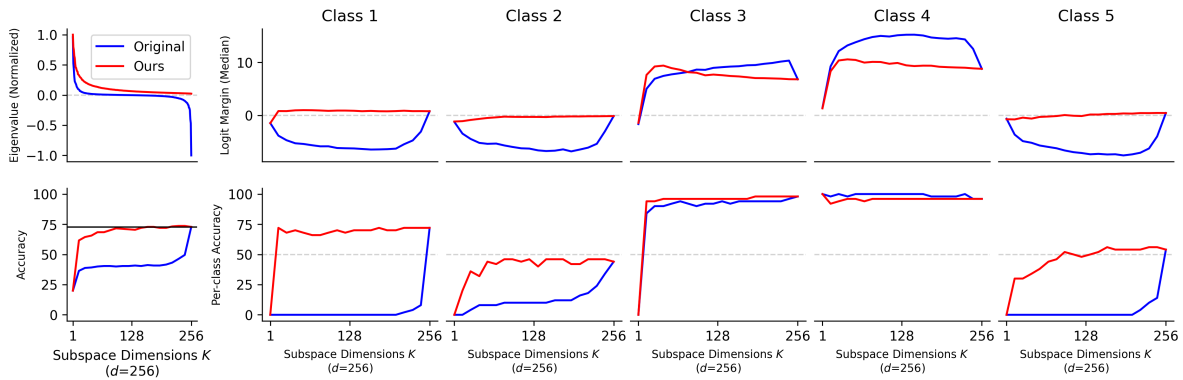

Figure C.2: Comparison between the behavior of the subspaces identified by the original PRCA [6] (blue) and ours (red), emphasizing that enforcing non-negativity is crucial for capturing subtask-relevant structure. The first column shows the eigenvalues of the two analyses' solution matrices (top) and the accuracy (bottom) of the ImageNet Domestic Cat subtask from ResNet18 from TorchVision, while the other columns show the median margin (Top) and per-class accuracy (Bottom) of each class. We compute the curves of each subspace by projecting the model's activations at Layer 3 onto the subspaces at various dimensionalities and track the corresponding statistics accordingly.

Fig. C.2 (First Column) shows in blue the relationship between the accuracy of the ImageNet Domestic Cat subtask from ResNet18 from TorchVision and the eigenvalues of the original PRCA's solution matrix $\mathbb{E}[\widetilde{a}_T c_T^\top + c_T \widetilde{a}_T^\top]$ [6]. From the two plots, we can see that the accuracy of the original

PRCA subspace gradually increases as we enlarge the subspace (by using more leading eigenvectors to construct the subspace). The accuracy, however, plateaus and only reaches the original value when incorporating the eigenvectors that correspond to the negative eigenvalues.

When inspecting per-class accuracy, we observe from the bottom row of Fig. C.2 that only classes 2 and 3 show a reasonable value, while the other classes have close to zero. Examining the top row, the margin between the target class's logit and the largest of the other classes, we see that these two classes have a substantially larger margin than the others. The evidence therefore suggests that the leading eigenvectors of the original PRCA's solution matrix tend to favor large-margin classes first. In contrast, as depicted with red in Fig. C.2, we see that the accuracy curve of our PRCA (which will show in Proposition 4 that it satisfies non-negativity) has a significantly better trend and reaches the original accuracy using only around 100 dimensions.

**Proposition 4** (Non-negativity). *Let $\langle \cdot, \cdot \rangle_U = \langle U^\top \cdot, U^\top \cdot \rangle$. The proposed objective ( Eq. (2) in the main text)*

$$\max_U \quad \left\{ \mathbb{E}\big[\langle \widetilde{a}, c \rangle_U \big] + \beta^{-1} \cdot \mathbb{E}\big[\langle \widetilde{a}, \widetilde{a} \rangle_U \big] + \beta \cdot \mathbb{E}\big[\langle c, c \rangle_U \big] \right\} \quad \text{s.t. } U^\top U = I_K$$

*is non-negative when $\beta > 0$.*

*Proof.* With $\Psi_\beta = \widetilde{a} c^\top + \beta^{-1} \widetilde{a} \widetilde{a}^\top + \beta c c^\top$ and $\langle x, y \rangle_U = x^\top U U^\top y$, the objective can be rewritten as $U^\top \mathbb{E}[\Psi_\beta] U$. Proving the non-negativity of the objective is then equivalent to showing that $v^\top \Psi_\beta v \geq 0$ for any $v \in \mathbb{R}^d$. Let $g = v^\top \widetilde{a}$ and $h = v^\top c$. We then have

$$v^\top \Psi_\beta v = gh + \beta^{-1} g^2 + \beta h^2$$
$$= \frac{1}{\beta}(g^2 + \beta g h) + \beta h^2$$

(using the identity $x^2 + bx = (x + \frac{b}{2})^2 - \frac{b^2}{4}$)

$$= \frac{1}{\beta}\left[\left(g + \frac{\beta h}{2}\right)^2 - \frac{\beta^2 h^2}{4}\right] + \beta h^2$$
$$= \frac{1}{\beta}\left(g + \frac{\beta h}{2}\right)^2 + \frac{3}{4}\beta h^2$$
$$\geq 0,$$

where the last step follows because $\beta > 0$. □

**Proposition 5** (Solution). *The solution of the proposed objective (Eq. (2) in the main text) is the eigenvectors of*

$$\Sigma_{ac} + 2\beta^{-1} \Sigma_a + 2\beta \Sigma_c,$$

*where $\Sigma_{ac} = \mathbb{E}[\widetilde{a} c^\top + c \widetilde{a}^\top]$, $\Sigma_a = \mathbb{E}[\widetilde{a} \widetilde{a}^\top]$, and $\mathbb{E}[c c^\top]$.*

*Proof.* We prove it by first showing the solution of the case $K = 1$ and then using the spectral theorem to extend it to any $K$.

We first substitute $u \in \mathbb{R}^d$ in the objective and construct the Lagrangian $\mathcal{L}$

$$\mathcal{L}(u) = u^\top \mathbb{E}[\widetilde{a}c^\top]u + \beta^{-1}u^\top \mathbb{E}[\widetilde{a}\widetilde{a}^\top]u + \beta u^\top \mathbb{E}[cc^\top]u - \lambda(u^\top u - 1).$$

Taking the derivative and setting it to zero implies

$$\left(\mathbb{E}[\widetilde{a}c^\top + c\widetilde{a}^\top] + 2\beta^{-1}\mathbb{E}[\widetilde{a}\widetilde{a}^\top] + 2\beta\mathbb{E}[cc^\top]\right)u = 2\lambda u.$$

Denoting LHS as $\Sigma_{ac}^+$, the equation above shows that $u$ is the eigenvector of $\Sigma_{ac}^+$. Given the symmetry of $\Sigma_{ac}^+$ and using the spectral theorem, the matrix $\Sigma_{ac}^+$ has orthogonal eigenvectors. Thus, these eigenvectors are the solution of the proposed objective.

□

**Supplementary Note D. Details of the Synthetic Experiment**

We recall that the goal of this experiment is to verify the benefits of our loss formulation on a small scale example using synthetic data. We assume that the student has no feedback beyond one layer-wise loss to isolate the effect of the output loss from the training.

In this supplementary note, we first describe how we synthesize the activations of the teacher and then outline the details of the student's architecture and training.

*Synthesizing Teacher Activations.* We generate the teacher's activations $a_T \in \mathbb{R}^d$ by transforming points from an interval to a higher-dimensional space using kernel PCA [7]. More concretely, we first choose $\Omega = \{\omega_i\}_{i=1}^d$ where $\omega \in [0, 2\pi)$ and compute the kernel matrix $T_\gamma \in \mathbb{R}^{d \times d}$ using the RBF kernel $\kappa_\gamma$ with scale $\gamma$. Let $Q$ and $\Lambda$ be the matrices containing the eigenvectors and eigenvalues of the kernel matrix $T_\gamma$. We define each activation vector $a_T \in \mathbb{R}^d$ to be the representation of some $\omega$ in the coordinates of the kernel PCA [8]

$$a_T = \Lambda^{-1/2} Q^\top \kappa_\Omega(\omega),$$

where the vector $\kappa_\Omega(\omega) = (\kappa_\gamma(\omega, \omega_i))_{i=1}^d$, and center these activations to have zero mean. We use $d = 2000$ and $\gamma = 5$. We choose the relevant region of the teacher's activations to be corresponding to $\omega \in [3\pi/5, 7\pi/5]$ and define the response vector $c_T$ of the activation $a_T$ to be itself if the corresponding $\omega$ is in the relevant interval and the zero vector otherwise.

*Student's Architecture and Training Details.* The architecture of the student is a one-layer neural network. More concretely, it takes $x \in \mathbb{R}^{d_0}$ to produce $a_\theta = \max(0, W_\theta x + b_\theta)$ where $W_\theta \in \mathbb{R}^{K \times d_0}$ and $b_\theta \in \mathbb{R}^K$ are the parameters of the student. For the input $x$, we synthesize it similar to $a_T$, except that we use a larger scale $\gamma' > \gamma$ and keep only the first $d_0$ dimensions of the kernel PCA. We construct the training set to be the vectors corresponding to a random subset $\Omega' \subset \Omega$.

We train the student to match its output $a_\theta$ with $a_T$ using the loss function $\mathcal{E}_l(a_T, a_\theta)$ of our formulation or the $(W, b)$-formulation. To make the learning dynamics of the two formulations comparable, we initialize $W_\theta$ and $W$ such that $\|W_\theta\|_F = \|W\|_F = \sqrt{K}$ and scale the student's bias $b_\theta$ proportionally; we set the bias of the $(W, b)$ formulation to be zero.

For the parameters, we use $\gamma' = 10$, $d_0 = 20$, $K = 5$, and $|\Omega'| = 500$ and perform the training for 250 epochs using SGD with mini-batches of size 32, momentum equal to 0.95, and a learning rate of 0.001. We repeat the training for five random initializations.

## Supplementary Note E. Details of Training and Explainable AI Analysis

*Training.* We provide the training details of experiments presented in the main text and the mapping between of the teacher's and the student's layers. We train the student with AdamW [9] for 100 epochs (starting with a learning rate at 0.001 and decaying every 25 epochs with a factor of 0.5) and use a batch size equal to 32. We use 20% of each subtask's original training instances as a validation set and use the accuracy on the set to determine early stopping. We repeat the training of each configuration for three random initializations. For the output-only distillation approach [10], we set the temperature value equal to 4 (similar to [11]) and also use weight decay, for which we use the validation set to select the penalty from $\{0, 10^{-5}, 10^{-4}, 10^{-3}, 10^{-2}\}$.

Table E.1 summarizes the mapping between the teacher and the student layers. When the spatial dimensions of the teacher's and the student's activations are different, we resize the larger one to make them compatible.

|  | Location | | | |
| --- | --- | --- | --- | --- |
| Teacher → Student | *1st* | *2nd* | *3rd* | *4th* |
| ResNet18 → ResNet18-S | `layer1` | `layer2` | `layer3` | `layer4` |
|  | *(same as above)* | | | |
| WResNet40 → ResNet18-S | `block1` `layer2` | `block2` `layer3` | `block3` `layer4` | × × |
| WResNet101 → MbNetv4 | `layer1` `blocks.1.1` | `layer2` `blocks.2.3` | `layer3` `blocks.3.1` | `layer4` `blocks.3.5` |
| ViTB16 → EffFormerv2 | `layers.2` `stages.0` | `layers.5` `stages.1` | `layers.8` `stages.2` | `layers.11` `stages.3` |

Table E.1: Correspondence between teacher and student layers from experiments presented in the main text. The symbol × indicates no mapping.

*Explainable AI.* We describe here the implementation details of our Explainable AI analysis presented in Section 4.2 of the main text as well as additional results. For each test instance, we compute a pixel-wise explanation associated to the model's prediction by attributing the difference between the target class's logit and the largest one of the other classes (Eq. (3) in the main text) to input pixels. We use Layer-wise Relevance Propagation (LRP) [12] as the primary attribution method in our analysis. To verify that our findings are not specific to LRP, we repeat the analysis of the 25% training data setup with Integrated Gradients (IntGrad) [13] and Shapley Value Sampling (SVS) [14].

We use the LRP implementation of Zennit [15] (with ResNetCanonizer and the EpsilonGammaBox composite) and set, for LRP-$\gamma$ [16] rule, $\gamma = 1$. For IntGrad and SVS, we use the implementations of Captum [17]. We perform IntGrad with 50 integration steps and a baseline value equal to zero. For SVS, we consider as input features patches of size $8 \times 8$, and execute SVS with 25 random permutations of features. We then calculate SVS attribution statistics from 10% of the test set.

## Supplementary Note F. Subtask-Relevant Subspace Estimation

We describe in this note the details on how we estimate our subtask-relevant subspace (discussed in Section 3.2). From Proposition 5, our subtask-relevant subspace can be constructed from the

leading eigenvectors of the following symmetric matrix

$$\Sigma_{ac} + 2\beta^{-1}\Sigma_a + 2\beta\Sigma_c,$$

where $\Sigma_{ac} = \mathbb{E}[\widetilde{a}_T c_T^\top + c_T \widetilde{a}_T^\top], \Sigma_a = \mathbb{E}[\widetilde{a}_T \widetilde{a}_T^\top]$, and $\Sigma_c = \mathbb{E}[c_T c_T^\top]$. Consequently, the estimation of the subspace reduces to estimating these three covariance matrices and solving the eigenvalue problem. We perform the estimation by collecting pairs of activations and their response vectors $(\widetilde{a}_T, c_T)$. To gather the pairs, we first, for each training instance of the subtask, take 20 activation vectors $a_T$ corresponding to some random spatial locations in the feature map at the chosen layer (or all if the number of locations is smaller); in the case of ViTB16, we equate its patch tokens to be the spatial locations. Next we compute $\mu_T = \mathbb{E}[a_T]$, $\widetilde{a}_T = a_T - \mu_T$, and

$$c_T = \frac{\partial \Delta}{\partial \widetilde{a}_T} = \frac{\partial \Delta}{\partial a_T}\frac{\partial a_T}{\partial \widetilde{a}_T} = \frac{\partial \Delta}{\partial a_T},$$

where $\Delta$ is defined as in Eq. (3) of the main text. We then construct the three covariance matrices from these pairs $(\widetilde{a}_T, c_T)$ using empirical averages and solve for the eigenvectors accordingly.

**Supplementary Note G. Additional Results**

*Small Training Size.* We report further results in Fig. G.3 for scenarios where we provide less training data (discussed in Section 4.1 of the main text). The results validate that the superiority of SubDistill is maintained in these lower data regimes.

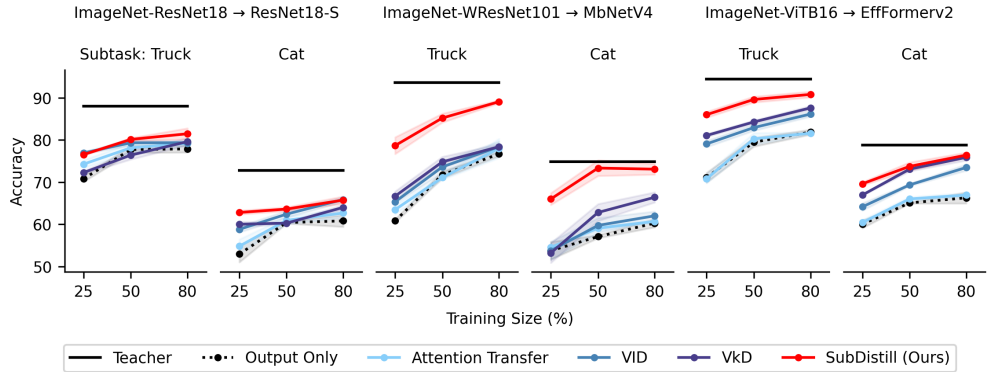

Figure G.3: Performance of students trained by with our SubDistill approach and other layer-wise distillation methods on the ImageNet subtasks using different amounts of training data. The shaded region indicates standard error from three random initializations. The results presented here complement the discussion in Section 4.1 of the main text.

*XAI Analysis.* We present here the replications of the same analysis presented in Section 4.2 on the other two ImageNet subtasks (namely, Truck and Domestic Cat) in Fig. G.5 and the results of the 25% training data setup using Integrated Gradients and Shapley Value Sampling. The observed trends, namely the rankings between methods, are consistent with correlation scores reported in the main paper.

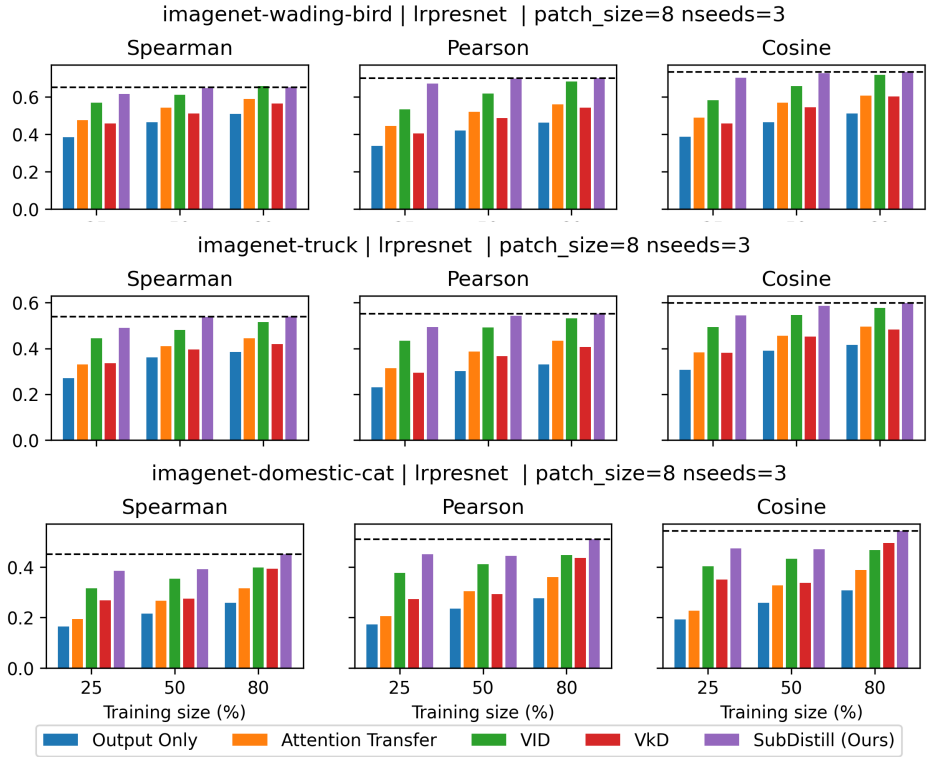

Figure G.4: Attribution statistics between student models from distilled models on three ImageNet subtasks and those of the teacher with attribution maps derived from LRP.

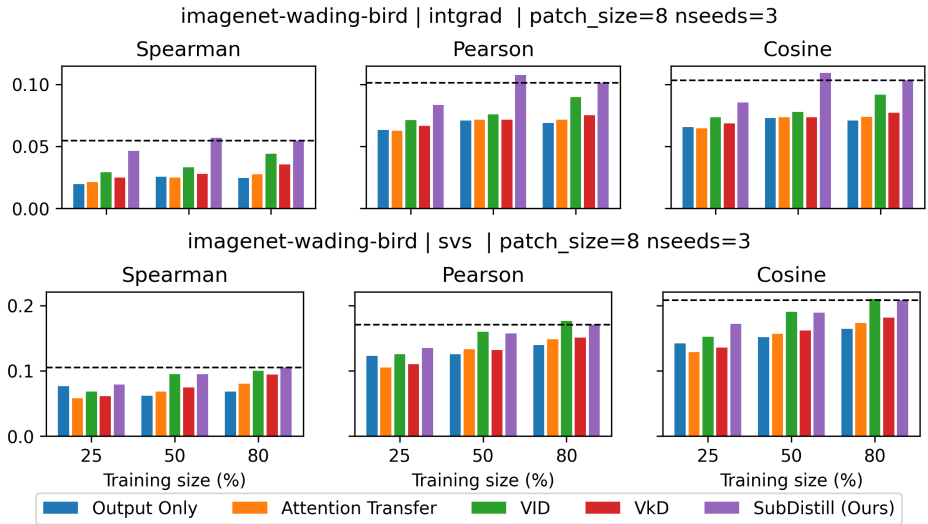

Figure G.5: Attribution statistics between student models from distilled models on the ImageNet 'wading bird' subtask and those of the teacher with attribution maps derived from Integrated Gradients (top) and Shapley Value Sampling (bottom).

9



\end{document}